\documentclass[a4paper]{article}
\usepackage{INTERSPEECH2021}
\usepackage{times}
\usepackage{epsfig}
\usepackage{graphicx}
\usepackage{amsmath}
\usepackage{amssymb}

\usepackage[breaklinks=true,bookmarks=false, colorlinks]{hyperref}
\usepackage{url}            
\usepackage{booktabs}       
\usepackage{amsfonts}       
\usepackage{nicefrac}       
\usepackage{microtype}      
\usepackage{graphicx}
\usepackage{amsmath}
\usepackage{subcaption}
\usepackage{textcomp}
\usepackage{multicol}
\usepackage{multirow}
\newcommand{\rarrow}{$\rightarrow$}

\usepackage{xcolor}
\usepackage{paralist}
\usepackage{cite}           

\title{Cascaded Multilingual Audio-Visual Learning from Videos}
\name{Andrew Rouditchenko$^1$, Angie Boggust$^1$, David Harwath$^2$, Samuel Thomas$^3$, Hilde Kuehne$^3$,\\ Brian Chen$^4$, Rameswar Panda$^3$, Rogerio Feris$^3$, Brian Kingsbury$^3$, Michael Picheny$^5$, James Glass$^1$}
\address{
  $^1$MIT CSAIL, USA \\ 
  $^2$UT Austin, USA \\
  $^3$IBM Research AI, USA \\
  $^4$Columbia University, USA\\
  $^5$NYU, USA}
\email{roudi@mit.edu}

\begin{document}

\maketitle
%

\begin{abstract}
In this paper, we explore self-supervised audio-visual models that learn from instructional videos.
Prior work has shown that these models can relate spoken words and sounds to visual content after training on a large-scale dataset of videos, but they were only trained and evaluated on videos in English.
To learn multilingual audio-visual representations, we propose a cascaded approach that leverages a model trained on English videos and applies it to audio-visual data in other languages, such as Japanese videos.
With our cascaded approach, we show an improvement in retrieval performance of nearly 10x compared to training on the Japanese videos solely.
We also apply the model trained on English videos to Japanese and Hindi spoken captions of images, achieving state-of-the-art performance.
\end{abstract}

\noindent\textbf{Index Terms}: multilingual, audio-visual, videos, self-supervised, low-resource

\section{Introduction}
\label{sec:introduction}

While technologies like Automatic Speech Recognition (ASR) and Machine Translation (MT) enable us to interact better with computers and each other, they are currently only available for less than 2\% of the world’s languages, in part due to the large amount of manually labelled data required for each language~\cite{prasad2019building}.
Recently, researchers have proposed models that can instead learn to recognize words from raw audio by associating them to semantically related images~\cite{synnaeve2014learning,harwath2016unsupervised,harwath2018jointly,harwath2020jointly,ilharco2019large,chrupala2017representations,merkx2019language}.
The first models were applied to English spoken audio captions, but further work applied the models to Hindi~\cite{harwath2018vision} and Japanese~\cite{havard2019models,ohishi2020trilingual} captions.
We are also interested in learning multilingual representations from audio-visual data, but we aim to do this from instructional videos that are naturally present on the internet and do not require recorded spoken captions.

To learn multilingual representations, we leverage the recently proposed Audio-Video Language Network~\cite{rouditchenko2020avlnet}.
Compared to prior image and spoken audio caption models, it learns from entire video clips and raw audio from instructional videos.
The model was trained on HowTo100M~\cite{miech2019howto100m}, a large-scale dataset of 1.2M instructional videos, and achieved strong video retrieval performance on the YouCook2~\cite{zhou2018towards} dataset of English cooking videos.

It would be challenging to collect large-scale instructional video datasets in other languages to train AVLnet given the significant engineering effort required to download and process them.
Furthermore, there are currently fewer instructional videos available for other languages, especially low-resource languages.
We address these limitations by proposing a cascaded approach that applies the AVLnet model trained on English videos to videos in Japanese.
While spoken audio captions of images already exist for Japanese~\cite{ohishi2020trilingual} and Hindi~\cite{harwath2018jointly}, there are no instructional video datasets similar in size to YouCook2 in other languages.
Therefore, we collected a dataset of instructional cooking videos in Japanese, named YouCook-Japanese.
Applying our cascaded approach, we show an improvement in retrieval performance of nearly 10x on YouCook-Japanese compared to training on the Japanese videos solely. 

\begin{figure}[t]
    \centering
    \includegraphics[width=0.9\linewidth]{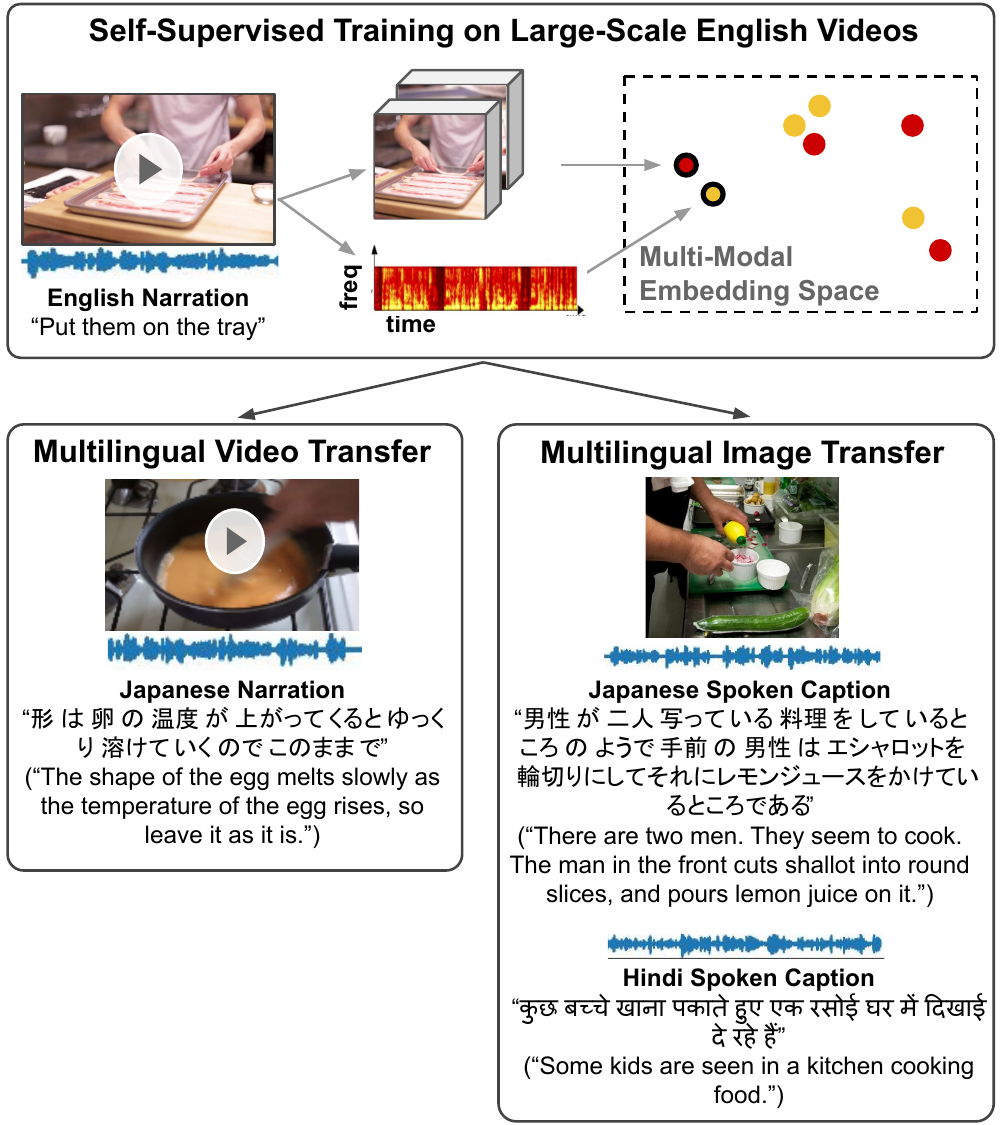}
    \caption{Given an audio-video model (AVLnet) trained on videos in English, we transfer the representations to videos in Japanese. We also transfer the representations to images and spoken captions in Japanese and Hindi.}
    \label{fig:overview}
\end{figure}
\begin{figure*}[t]
    \centering 
    \begin{subfigure}{0.9\linewidth}
        \includegraphics[width=\textwidth]{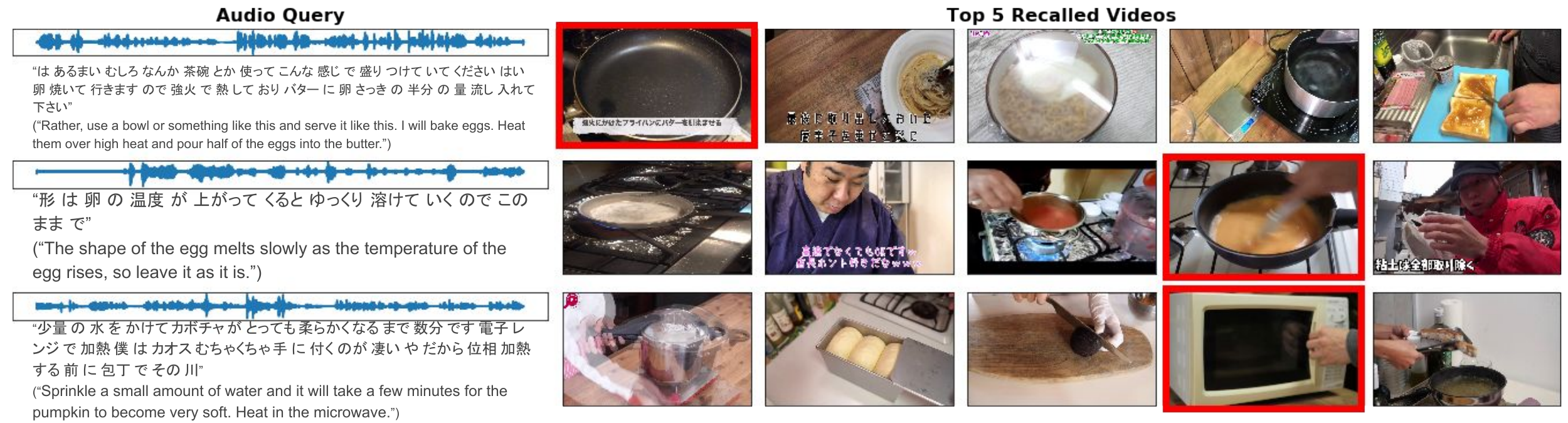}
    \end{subfigure}
    \vfill
    \begin{subfigure}{0.9\linewidth}
        \includegraphics[width=\textwidth]{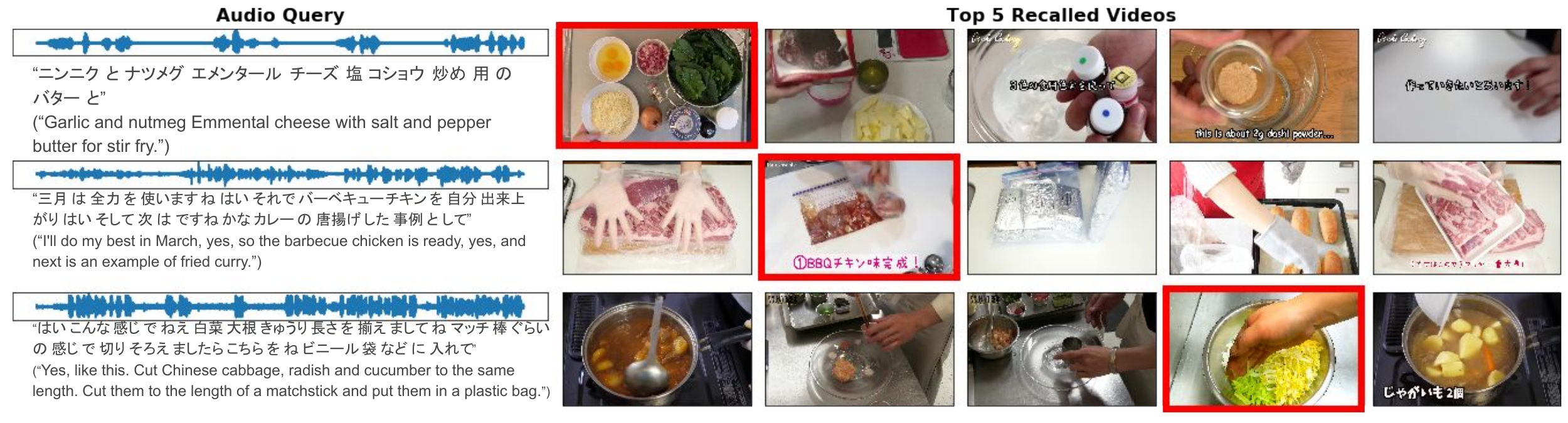}
    \end{subfigure}
    \caption{YouCook-Japanese video retrieval results with AVLnet. Top: retrieval after  training on HowTo100M and without fine-tuning on YouCook-Japanese (zero-shot). Bottom: retrieval after  training on HowTo100M and fine-tuning on YouCook-Japanese. Japanese ASR transcripts and English translations are shown, but AVLnet only uses audio as input. The correct clip match is in red.}
    \label{fig:retrieval}
\end{figure*}
We also show that our cascaded approach can work as a bridge between English instructional videos and the spoken audio captions of images in Japanese and Hindi.
Given the AVLnet model trained on English videos, we fine-tune it on Japanese and Hindi spoken captions of images, achieving state-of-the-art performance.
Finally, we provide an analysis of the impact of the amount of English training videos on retrieval performance.
We will release our code, trained models, and data at \href{http://avlnet.csail.mit.edu}{http://avlnet.csail.mit.edu}

\section{Related Work}
\label{sec:related-work}

\noindent \textbf{Image and Spoken Caption Models.}
Several works~\cite{synnaeve2014learning,harwath2015deep,harwath2016unsupervised} demonstrate the ability to learn semantic relationships between objects in images and the spoken words describing them using only the pairing between images and spoken captions as supervision.
Using this framework, researchers have proposed improved image encoders, audio encoders, and loss functions~\cite{harwath2018jointly,harwath2020learning,harwath2020jointly,chrupala2017representations,merkx2019language,ilharco2019large,suris2019learning,mortazavi2020speech,sanabria2021talk,wang2021align}.
Harwath~et~al.~\cite{harwath2016unsupervised,harwath2017learning,harwath2018jointly} collected 400k spoken audio captions of images in the Places205~\cite{zhou2014learning} dataset in English, which is one of the largest spoken caption datasets.
For a recent survey of visually grounded models of spoken language, see Chrupała~\cite{chrupala2021visually}.
We instead use videos already available on the internet in English and other languages as our main source of training data.
While we use the spoken narration naturally present in instructional videos, researchers have recently collected spoken captions for videos~\cite{monfort2021spoken,oncescu2020queryd}.

\noindent \textbf{Audio-Video Models.}
Instructional videos serve as a rich source of training data for learning the relationships between spoken words and visual content.
While several models~\cite{holzenberger2019learning,wray2019fine,alayrac2020self,chen2021multimodal} have been proposed to learn from large-scale datasets such as How2~\cite{sanabria18how2} and HowTo100M~\cite{miech2019howto100m}, they rely on having text transcripts.
Our work instead builds from models that learn from the raw visual and audio channels.
Boggust et al.~\cite{boggust2019grounding} applied an image-caption~\cite{harwath2018jointly} model to videos, using a single image frame from entire video clips to perform video to audio retrieval.
Rouditchenko et al.~\cite{rouditchenko2020avlnet} proposed the AVLnet model that pools visual information over entire clips using 2D and 3D visual CNNs.
We use AVLnet trained on English HowTo100M videos and apply it to cooking videos in Japanese and images and spoken captions in Japanese and Hindi.

\noindent \textbf{Multilingual Speech and Video Processing.}
The image and spoken caption models have been explored in the multilingual setting.
Harwath et al.~\cite{harwath2018vision} collected 100k Hindi captions of Places images and proposed a bilingual audio-visual model.
Building from this, Ohishi et al.~\cite{ohishi2020trilingual} collected 100k Japanese captions and proposed a trilingual model.
Other work has proposed bilingual models with synthetic spoken captions~\cite{havard2019models} and image text taggers~\cite{kamper2018visually}, clustering for bilingual image-audio dictionaries~\cite{azuh2019towards}, and pair expansion methods for learning from multilingual captions of disjoint images~\cite{ohishi2020pair}.
Instead of learning from multiple languages simultaneously, our approach is to learn from them one at a time in a cascade.

Several multilingual video datasets have been introduced, such as How2~\cite{sanabria18how2} and VATEX~\cite{wang2019vatex} which contain parallel translations of English video captions in Portuguese and Chinese.
Instead of collecting parallel translations, Sigurdsson et al. proposed versions of HowTo100M in Japanese, French, and Korean~\cite{sigurdsson2020visual}.
Thus far, all of the methods proposed on these datasets rely on text captions.
Instead, we use AVLnet to learn from videos using speech audio and without requiring text.

Finally, multilingual ASR methods include simultaneous training on multiple languages~\cite{huang2013cross,karafidt2018analysis,cho2018multilingual,conneau2020unsupervised} and cascaded approaches in which representations learned from one language are used as initialization for other languages~\cite{thomas2012multilingual,ghoshal2013multilingual}.
Our approach is similar to the cascaded methods, but it only requires audio-visual data without transcripts.
\section{Technical Approach}
\label{sec:methods}

\subsection{Videos}
Our goal is to learn audio-visual representations for videos in languages other than English using AVLnet~\cite{rouditchenko2020avlnet}.
AVLnet is trained through a contrastive loss to discriminate between temporally aligned audio-video pairs and temporally mismatched pairs from both within the same video and from other videos. 
This results in an audio-video embedding space which colocates semantically similar audio and visual inputs.
Since AVLnet does not require any annotations besides the raw video data, we only assume that a set of videos in the target language is given, but without any additional annotation.
One approach is to simply train AVLnet only on the target videos in the new language.
However, we find that hundreds of thousands of videos are necessary to learn strong representations, and there is simply not enough videos in datasets such as YouCook2 to train the model from scratch.
Therefore, our proposed approach is simple: given the AVLnet model trained on English HowTo100M videos, we apply it to videos in Japanese by directly fine-tuning it on the Japanese videos.
This represents a cascade since the model only learns from videos in one language at a time (ie. first English, then Japanese).

\label{sec:methods-video-transfer}
\noindent \textbf{YouCook-Japanese.}
There are currently no other instructional video datasets in other languages similar in size to YouCook2.
Therefore, we collected a dataset of Japanese cooking videos, and call it YouCook-Japanese to indicate the similarity in content and size to YouCook2.
As a starting point, Sigurdsson et al.~\cite{sigurdsson2020visual} proposed a version of HowTo100M in Japanese with approximately 300k videos. 
We followed the steps to download Japanese instructional videos from YouTube, except we limited the search to cooking videos only.
We used a CNN-based audio segmentation toolkit~\cite{ddoukhanicassp2018} to segment the videos into clips containing speech, and then filtered the clips to be at least 5s and at most 50s.
To make the dataset similar in size to YouCook2, we originally selected 10k random clips for training, 3k clips for validation, and 3k clips for evaluation, with the constraint that each video can only appear in one set.
We then manually inspected the videos and removed those unrelated to cooking and which did not contain Japanese audio.
In the end, the training set contains 7,777 clips from 608 videos, the validation set contains 2,156 clips from 179 videos, and the evaluation set contains 2,128 clips from 162 videos.
We report results on the held-out evaluation set and encourage other researchers to tune parameters only on the validation set.

\begin{table}[t]
    \begin{center}
    \caption{Video retrieval on YouCook2 Videos (YC-EN) and YouCook-Japanese videos (YC-JP). HT100M=HowTo100M.}
    \label{tab:video}
    \vspace{-0.30cm}
    \begin{subtable}{\linewidth}
         \caption{English YouCook2 Videos (YC-EN)}
         \vspace{-0.20cm}
        \resizebox{\linewidth}{!}{\begin{tabular}{lrrrrrr}
        \toprule
        \multicolumn{1}{l}{\multirow{2}{*}{\textbf{AVLnet Train Data}}} & \multicolumn{3}{c}{\textbf{Video Clip  (A\rarrow V)}}    & \multicolumn{3}{c}{\textbf{Language  (V\rarrow A)}}\\
        \multicolumn{1}{c}{}   & \multicolumn{1}{c}{R@1}  & \multicolumn{1}{c}{R@5}  & \multicolumn{1}{c}{R@10} & \multicolumn{1}{c}{R@1}  & \multicolumn{1}{c}{R@5}  & \multicolumn{1}{c}{R@10} \\
        \midrule
        Random & 0.03 & 0.15 & 0.3 & 0.03 & 0.15 & 0.3  \\
        \midrule
        YC-EN & 0.7 & 2.3 & 3.9 & 0.8 & 3.0 & 4.9 \\  
        HT100M & 27.4 & 51.6 & 61.5 & 27.3 & 51.2 & 60.8 \\  
        HT100M + YC-EN & \textbf{30.7} & \textbf{57.7} & \textbf{67.4} & \textbf{33.0} & \textbf{58.9} & \textbf{68.4} \\
        HT100M + YC-JP & 18.8 & 40.5 & 51.2 & 20.8 & 44.3 & 53.3 \\
        \bottomrule
        \end{tabular}}
        \vspace{0.15cm}
        \label{tab:video-english}
    \end{subtable}
    
    
    \begin{subtable}{\linewidth}
        \caption{YouCook-Japanese Videos (YC-JP)}
        \vspace{-0.20cm}
        \resizebox{\linewidth}{!}{
        \begin{tabular}{lrrrrrr}
        \toprule
        \multicolumn{1}{l}{\multirow{2}{*}{\textbf{AVLnet Train Data}}} & \multicolumn{3}{c}{\textbf{Video Clip (A\rarrow V)}}    & \multicolumn{3}{c}{\textbf{Language (V\rarrow A)}}\\
        \multicolumn{1}{c}{}    & \multicolumn{1}{c}{R@1}  & \multicolumn{1}{c}{R@5}  & \multicolumn{1}{c}{R@10} & \multicolumn{1}{c}{R@1}  & \multicolumn{1}{c}{R@5}  & \multicolumn{1}{c}{R@10} \\
        \midrule
        Random & 0.03 & 0.17 & 0.33 & 0.03 & 0.17 & 0.33  \\
        \midrule
        YC-JP & 0.6 & 2.2 & 4.1 & 0.4 & 1.7 & 3.5 \\  
        HT100M & 4.4 & 11.9 & 18.2 & 6.0 & 14.9 & 21.9 \\  
        HT100M + YC-EN & 4.1 & 11.6 & 17.8 & 5.8 & 14.8 & 20.6 \\
        HT100M + YC-JP & \textbf{7.8} & \textbf{22.8} & \textbf{32.1} & \textbf{9.0} & \textbf{23.3} & \textbf{33.1} \\
        \bottomrule
        \end{tabular}}
        \label{tab:video-japanese}
    \end{subtable}
\end{center}
\vspace{-0.3cm}
\end{table}

\subsection{Images and Spoken Captions}
\label{sec:methods-image-transfer}
Since instructional videos and spoken captions of images both contain descriptive audio of visual scenes, our cascaded approach is also applicable to images and spoken captions.
Specifically, we use the AVLnet model trained on HowTo100M videos and fine-tune it on the spoken captions and images in the Places Audio Caption Dataset in Japanese and Hindi.
For these experiments, we train AVLnet using only the 2D features in the visual branch so that the model can work on both videos and images.

\begin{table}[t]
    \begin{center}
    \caption{Image retrieval on the Places Audio Caption dataset.}
    \label{tab:places}
    \vspace{-0.30cm}
    \begin{subtable}{\linewidth}
        \caption{Places Audio Captions - Japanese}
        \vspace{-0.20cm}
        \resizebox{\linewidth}{!}{\begin{tabular}{llccccc}
        \toprule
        \multicolumn{1}{l}{\multirow{2}{*}{\textbf{Method}}} & \multicolumn{3}{c}{\textbf{Audio to Image}} & \multicolumn{3}{c}{\textbf{Image to Audio}}\\
        \multicolumn{1}{c}{}    & \multicolumn{1}{c}{R@1}  & \multicolumn{1}{c}{R@5}  & \multicolumn{1}{c}{R@10} & \multicolumn{1}{c}{R@1}  & \multicolumn{1}{c}{R@5}  & \multicolumn{1}{c}{R@10} \\
        \midrule
        Random & 0.1 & 0.5 & 1.0 & 0.1 & 0.5 & 1.0 \\
        Havard et al. \cite{havard2019models} & 18.2 & 48.5 & 62.2 & 15.3 & 41.4 & 57.6  \\
        Ohishi et al. \cite{ohishi2020pair} & 20.1 & 49.7 & 63.9 & 16.7 & 44.3 & 57.8  \\
        Ohishi et al. \cite{ohishi2020trilingual} & 20.3 & 52.0 & 66.7 & 20.0 & 46.8 & 62.3  \\
        \textbf{AVLnet} & \textbf{23.5} & \textbf{57.3} & \textbf{70.4} & \textbf{24.3} & \textbf{56.6} & \textbf{70.0} \\
        \bottomrule
        \end{tabular}}
        \label{tab:places-japanese}
    \end{subtable}

    \begin{subtable}{\linewidth}
        \vspace{0.15cm}
        \caption{Places Audio Captions - Hindi}
        \vspace{-0.20cm}
        \resizebox{\linewidth}{!}{\begin{tabular}{llccccc}
        \toprule
        \multicolumn{1}{l}{\multirow{2}{*}{\textbf{Method}}} & \multicolumn{3}{c}{\textbf{Audio to Image}} & \multicolumn{3}{c}{\textbf{Image to Audio}}\\
        \multicolumn{1}{c}{}    & \multicolumn{1}{c}{R@1}  & \multicolumn{1}{c}{R@5}  & \multicolumn{1}{c}{R@10} & \multicolumn{1}{c}{R@1}  & \multicolumn{1}{c}{R@5}  & \multicolumn{1}{c}{R@10} \\
        \midrule
        Random & 0.1 & 0.5 & 1.0 & 0.1 & 0.5 & 1.0 \\
        Harwath et al. \cite{harwath2018vision} & 8.0 & 25.0 & 35.6 & 7.4 & 23.5 & 35.4  \\
        Havard et al. \cite{havard2019models} & 9.6 & 28.2 & 40.7 & 8.0 & 27.6 & 37.1  \\
        Ohishi et al. \cite{ohishi2020pair} & 9.4 & 29.8 & 41.8 & 9.3 & 29.5 & 38.2  \\
        Ohishi et al. \cite{ohishi2020trilingual} & 11.2 & 31.5 & 44.5 & 10.8 & 31.3 & 41.9  \\
        \textbf{AVLnet} & \textbf{15.2} & \textbf{38.9} & \textbf{51.1} & \textbf{17.0} & \textbf{39.8} & \textbf{51.5} \\
        \bottomrule
        \label{tab:places-hindi}
        \end{tabular}}
    \end{subtable}
    \end{center}
    \vspace{-0.3cm}
\end{table}

\begin{table}[t]
    \begin{center}
    \caption{Comparison of frozen versus trainable image encoder for fine-tuning on the Places Audio Caption dataset.}
   \vspace{-0.3cm}
    \label{tab:ablation}
    \begin{subtable}{\linewidth}
        \resizebox{\linewidth}{!}{\begin{tabular}{llcccccc}
        \toprule
        \multicolumn{1}{l}{\multirow{2}{*}{\textbf{Language}}} & \textbf{Frozen} & \multicolumn{3}{c}{\textbf{Audio to Image}} & \multicolumn{3}{c}{\textbf{Image to Audio}}\\
        \multicolumn{1}{c}{} & \textbf{Img. CNN} & \multicolumn{1}{c}{R@1}  & \multicolumn{1}{c}{R@5}  & \multicolumn{1}{c}{R@10} & \multicolumn{1}{c}{R@1}  & \multicolumn{1}{c}{R@5}  & \multicolumn{1}{c}{R@10} \\
        \midrule
        \multirow{2}{*}{\shortstack[l]{Japanese}} & \textbf{Yes} & \textbf{23.5} & \textbf{57.3} & \textbf{70.4} & \textbf{24.3} & \textbf{56.6} & \textbf{70.0}  \\
        & No & 20.8 & 50.9 & 64.9 & 20.9 & 49.5 & 63.5   \\
        \midrule
        \multirow{2}{*}{\shortstack[l]{Hindi}} & \textbf{Yes} & \textbf{15.2} & \textbf{38.9} & \textbf{51.1} & \textbf{17.0} & \textbf{39.8} & \textbf{51.5}  \\
        & No & 12.1 & 30.9 & 44.1 & 11.9 & 30.8 & 41.7  \\
        \bottomrule
        \end{tabular}}
    \end{subtable}
    \end{center}
   \vspace{-0.3cm}
\end{table}
\section{Experiments}
\label{sec:experiments}

\subsection{Datasets}
\label{sec:experiments-datasets}
\noindent \textbf{Videos.}
We use the following instructional video datasets: HowTo100M~\cite{miech2019howto100m} (1.2M videos), YouCook2~\cite{zhou2018towards} and YouCook-Japanese.
For YouCook2, we use 9,586 train clips and 3,350 validation clips as in \cite{rouditchenko2020avlnet}.
We evaluate performance on audio to video clip retrieval and video clip to audio retrieval using the standard recall metrics R@1, R@5, R@10.

\noindent \textbf{Images and Spoken Captions.}
We fine-tune and evaluate our model on the Places Audio Caption dataset~\cite{harwath2018jointly}, which contains 100k images from the Places205 dataset~\cite{zhou2014learning} each with a spoken caption in Japanese~\cite{ohishi2020trilingual} and Hindi~\cite{harwath2018vision}.
We evaluate the performance on audio to image and image to audio retrieval using the standard recall metrics R@1, R@5, R@10.
We follow the prior work~\cite{harwath2018vision,ohishi2020trilingual} and report results on the validation sets of 1k images and spoken captions.

\subsection{Implementation Details}
\label{sec:experiments-implementation}
For training AVLnet on HowTo100M, we follow the details in \cite{rouditchenko2020avlnet}.
Each batch contains $128$ videos with $M=32$ clips per video, where each clip is $t=10$ seconds long, for an effective batch size of $4,096$ clips per batch.
The 2D visual feature extractor is a pre-trained ResNet-152~\cite{he2016deep} model and the 3D visual feature extractor is a pre-trained ResNeXt-101~\cite{hara2018can} model.
Both are kept frozen during training. 
The trainable audio encoder is the ResDAVEnet model ~\cite{harwath2020jointly} which operates on log Mel filterbank spectrograms.
For fine-tuning on video clips from YouCook2 and YouCook-Japanese, we use a batch size of 256 clips and a learning rate of $1\mathrm{e}{-4}$.
We pad the audio or crop it up to 50 seconds in length.
For fine-tuning AVLnet on images and spoken captions in Places, we either keep the ResNet-152 model frozen or fine-tune it.
We use a learning rate of $1\mathrm{e}{-3}$ for the frozen setting and a learning rate of $1\mathrm{e}{-4}$ for the trainable setting.
Models were trained with the MMS loss~\cite{ilharco2019large}.

\begin{figure}[t]
    \centering 
    \begin{subfigure}{0.85\linewidth}
        \includegraphics[width=\textwidth]{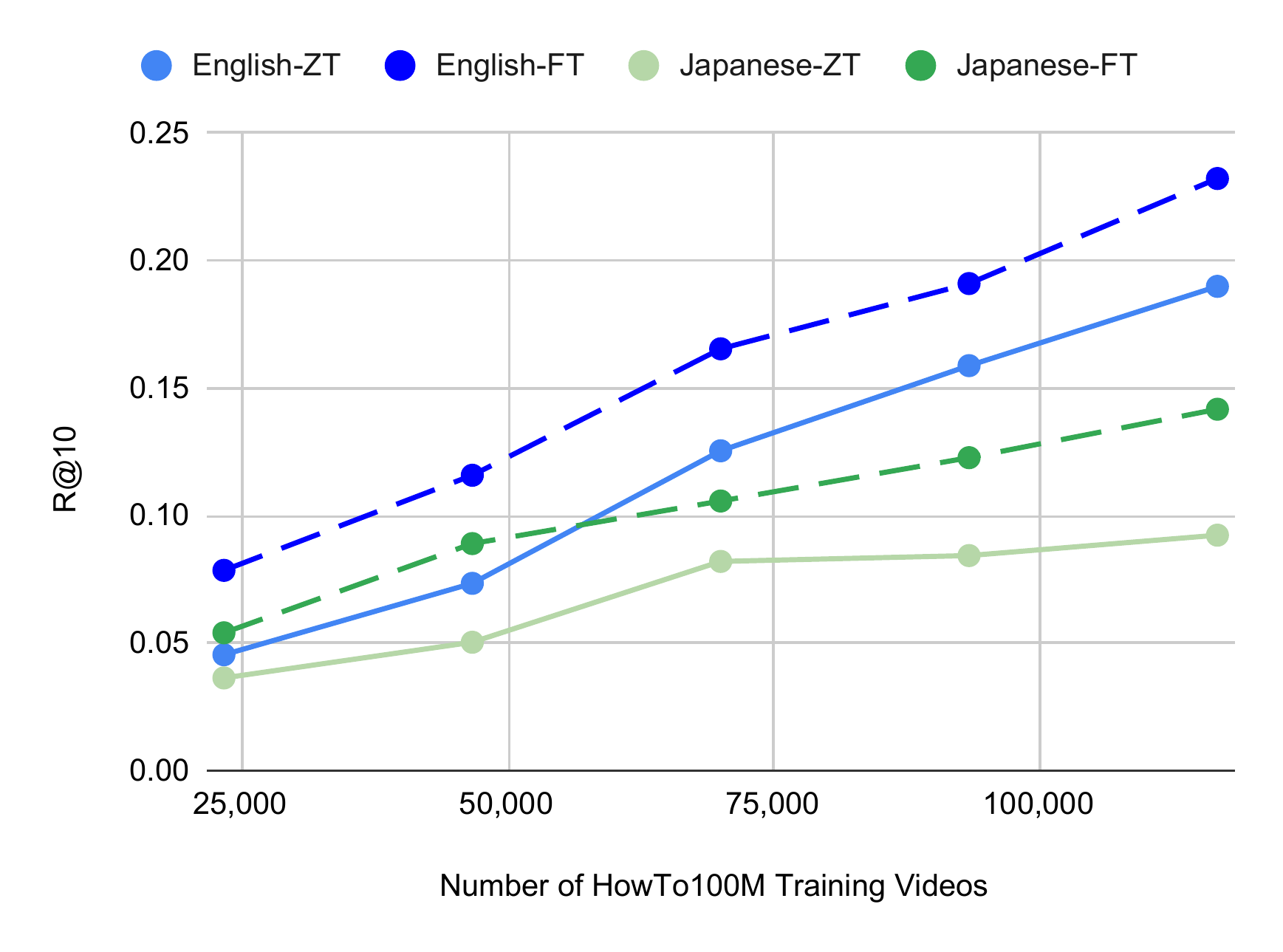}
    \end{subfigure}
    \vfill
    \begin{subfigure}{0.85\linewidth}
        \includegraphics[width=\textwidth]{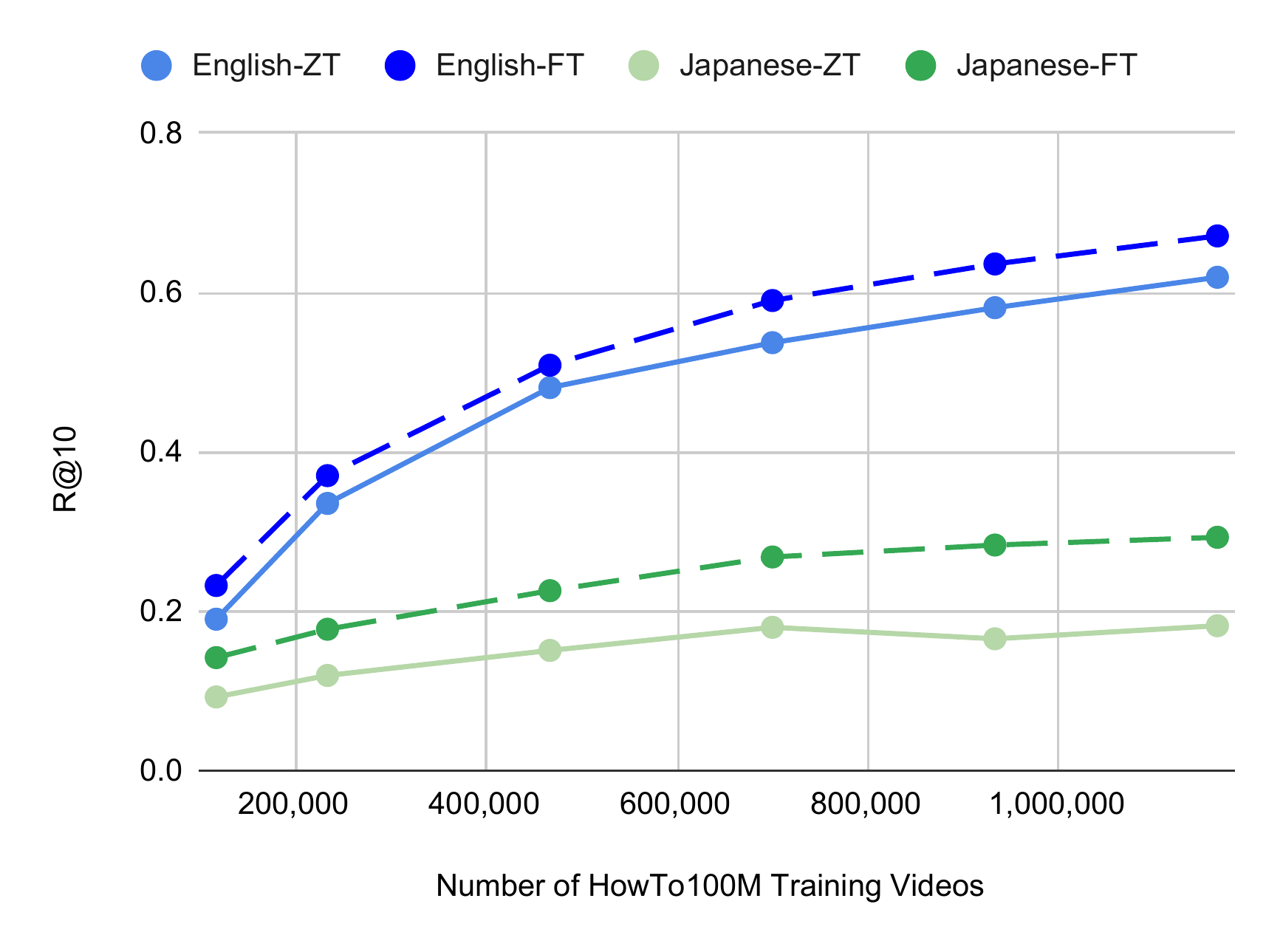}
    \end{subfigure}
    \caption{Video retrieval performance when varying the \% of HowTo100M videos. Top: \{2,4,6,8,10\}\%. Bottom: \{10,20,40,60,80,100\}\%. ZT=Zero-Shot, FT=Fine-tune.}
    \label{fig:transfer}
\end{figure}

\subsection{Video Retrieval}
\label{sec:experiments-video}
\noindent \textbf{YouCook2.} 
Table~\ref{tab:video-english} shows the video retrieval results on English YouCook2 videos. 
We note that some of the YouCook2 results have already been presented in~\cite{rouditchenko2020avlnet}, and we re-print them here for comparison with the results on YouCook-Japanese.
Training on HowTo100M significantly improves performance compared with training only on YouCook2.
In the zero-shot setting, ie. without fine-tuning on any YouCook2 videos, the model achieves strong retrieval performance, likely due to the similar instructional domain of HowTo100M and YouCook2 and shared language (English).
Performance further improves after fine-tuning on YouCook2 videos.
The final row of Table~\ref{tab:video-english} shows that fine-tuning the model on YouCook-Japanese videos reduces the performance on YouCook2, indicating that the model is sensitive to the language present in the videos.

\noindent \textbf{YouCook-Japanese.}
Table~\ref{tab:video-japanese} shows the video retrieval results on YouCook-Japanese videos.
AVLnet's performance when trained only on YouCook-Japanese is similar to AVLnet's performance on YouCook2 when trained only on YouCook2 videos, indicating that the two datasets are similar in difficulty.
Using our cascaded approach, we apply the AVLnet model trained on HowTo100M to the Japanese videos which significantly improves performance.
In the zero-shot setting, ie. without fine-tuning, the retrieval performance is nearly 5x the performance compared with training on YouCook-Japanese only.
This is surprising considering that the model has only been trained on English videos.
Fine-tuning the model on the Japanese videos further increases the performance to nearly 10x the performance compared with training on YouCook-Japanese only.
We also note that fine-tuning the model on English YouCook2 videos instead of Japanese videos is comparable to the zero-shot performance, further indicating that the model is actually sensitive to the language present in the videos.

\noindent \textbf{Qualitative results.}
Figure~\ref{fig:retrieval} shows qualitative YouCook-Japanese video retrieval results.
In the zero-shot setting, without fine-tuning on Japanese videos, the model seems to perform retrieval using salient natural sounds, for example, sizzling sounds or microwave beeps.
After fine-tuning the model on YouCook-Japanese, the model can handle more complex queries and retrieve video clips with specific ingredients mentioned in the audio queries.

\noindent \textbf{Varying the \% of HowTo100M videos.}
In Figure~\ref{fig:transfer}, we show the video retrieval performance when training AVLnet with a smaller percentage of HowTo100M videos.
The plots show that performance generally increases with the number of HowTo100M videos.
The gap between performance on English and Japanese videos is lower with 10\% or less of the HowTo100M videos.

\subsection{Image Retrieval}
\label{sec:experiments-image}
Table~\ref{tab:places} shows the retrieval results on the Places Audio Caption dataset in Hindi and Japanese.
For our cascaded approach, we fine-tune AVLnet trained on HowTo100M videos to each language in Places independently. 
We compare our approach to the state-of-the-art models for each dataset.
While previous models are not trained on HowTo100M videos, some of them~\cite{harwath2018vision,ohishi2020trilingual} are trained on images with parallel spoken captions in multiple languages.
Our cascaded approach involves training on one language at a time, achieving large gains over prior baselines.

Table~\ref{tab:ablation} shows that retrieval results were higher for Japanese and Hindi with a frozen visual encoder (ResNet-152) than with a trainable encoder.
We hypothesize that 100k images in the Japanese and Hindi training set is not enough to fine-tune the ResNet-152, and therefore it is better to leave it frozen.
Furthermore, given that the visual encoders are also frozen during fine-tuning on videos, these results suggest that the visual branch is more language independent than the audio branch, and that the audio branch needs to be adapted to handle unseen languages.

\section{Conclusion}
\label{sec:conclusion}
We propose a cascaded approach to learn multilingual audio-visual representations.
Given the AVLnet model trained on English HowTo100M videos, we fine-tuned and evaluated it on YouCook-Japanese videos and the images and spoken captions in the Places Audio Caption dataset in Japanese and Hindi.
Our cascaded improves performance on the Japanese videos by nearly 10x compared to training on the Japanese videos solely.
Our approach could plausibly work for instructional videos in any language.
One direction that we plan to explore in the future is cross-lingual retrieval in videos, for example, retrieving Japanese audio from English audio through related videos.

\section{Acknowledgements}
This research was supported by the MIT-IBM Watson AI Lab. We thank the IBM Japan team for help with Japanese ASR.
\bibliographystyle{IEEEtran}
\bibliography{avlnet}

\begin{thebibliography}{10}
\providecommand{\url}[1]{#1}
\csname url@samestyle\endcsname
\providecommand{\newblock}{\relax}
\providecommand{\bibinfo}[2]{#2}
\providecommand{\BIBentrySTDinterwordspacing}{\spaceskip=0pt\relax}
\providecommand{\BIBentryALTinterwordstretchfactor}{4}
\providecommand{\BIBentryALTinterwordspacing}{\spaceskip=\fontdimen2\font plus
\BIBentryALTinterwordstretchfactor\fontdimen3\font minus
  \fontdimen4\font\relax}
\providecommand{\BIBforeignlanguage}[2]{{%
\expandafter\ifx\csname l@#1\endcsname\relax
\typeout{** WARNING: IEEEtran.bst: No hyphenation pattern has been}%
\typeout{** loaded for the language `#1'. Using the pattern for}%
\typeout{** the default language instead.}%
\else
\language=\csname l@#1\endcsname
\fi
#2}}
\providecommand{\BIBdecl}{\relax}
\BIBdecl

\bibitem{prasad2019building}
M.~Prasad, D.~van Esch, S.~Ritchie, and J.~F. Mortensen, ``Building
  large-vocabulary asr systems for languages without any audio training data.''
  in \emph{INTERSPEECH}, 2019.

\bibitem{synnaeve2014learning}
G.~Synnaeve, M.~Versteegh, and E.~Dupoux, ``Learning words from images and
  speech,'' \emph{NeurIPS Workshop on Learning Semantics}, 2014.

\bibitem{harwath2016unsupervised}
D.~Harwath, A.~Torralba, and J.~Glass, ``Unsupervised learning of spoken
  language with visual context,'' in \emph{NeurIPS}, 2016.

\bibitem{harwath2018jointly}
D.~Harwath, A.~Recasens, D.~Sur{\'\i}s, G.~Chuang, A.~Torralba, and J.~Glass,
  ``Jointly discovering visual objects and spoken words from raw sensory
  input,'' in \emph{ECCV}, 2018.

\bibitem{harwath2020jointly}
------, ``Jointly discovering visual objects and spoken words from raw sensory
  input,'' \emph{IJCV}, 2020.

\bibitem{ilharco2019large}
G.~Ilharco, Y.~Zhang, and J.~Baldridge, ``Large-scale representation learning
  from visually grounded untranscribed speech,'' in \emph{CoNLL}, 2019.

\bibitem{chrupala2017representations}
G.~Chrupa{\l}a, L.~Gelderloos, and A.~Alishahi, ``Representations of language
  in a model of visually grounded speech signal,'' in \emph{ACL}, 2017.

\bibitem{merkx2019language}
D.~Merkx, S.~L. Frank, and M.~Ernestus, ``Language learning using speech to
  image retrieval,'' in \emph{INTERSPEECH}, 2019.

\bibitem{harwath2018vision}
D.~Harwath, G.~Chuang, and J.~Glass, ``Vision as an interlingua: Learning
  multilingual semantic embeddings of untranscribed speech,'' in \emph{ICASSP},
  2018.

\bibitem{havard2019models}
W.~N. Havard, J.-P. Chevrot, and L.~Besacier, ``Models of visually grounded
  speech signal pay attention to nouns: A bilingual experiment on english and
  japanese,'' in \emph{ICASSP}, 2019.

\bibitem{ohishi2020trilingual}
Y.~Ohishi, A.~Kimura, T.~Kawanishi, K.~Kashino, D.~Harwath, and J.~Glass,
  ``Trilingual semantic embeddings of visually grounded speech with
  self-attention mechanisms,'' in \emph{ICASSP}, 2020.

\bibitem{rouditchenko2020avlnet}
A.~Rouditchenko, A.~Boggust, D.~Harwath, B.~Chen, D.~Joshi, S.~Thomas,
  K.~Audhkhasi, H.~Kuehne, R.~Panda, R.~Feris \emph{et~al.}, ``Avlnet: Learning
  audio-visual language representations from instructional videos,''
  \emph{arXiv preprint arXiv:2006.09199}, 2020.

\bibitem{miech2019howto100m}
A.~Miech, D.~Zhukov, J.-B. Alayrac, M.~Tapaswi, I.~Laptev, and J.~Sivic,
  ``Howto100m: Learning a text-video embedding by watching hundred million
  narrated video clips,'' in \emph{ICCV}, 2019.

\bibitem{zhou2018towards}
L.~Zhou, C.~Xu, and J.~J. Corso, ``Towards automatic learning of procedures
  from web instructional videos,'' in \emph{AAAI}, 2018.

\bibitem{harwath2015deep}
D.~Harwath and J.~Glass, ``Deep multimodal semantic embeddings for speech and
  images,'' in \emph{ASRU}, 2015.

\bibitem{harwath2020learning}
D.~Harwath, W.-N. Hsu, and J.~Glass, ``Learning hierarchical discrete
  linguistic units from visually-grounded speech,'' in \emph{ICLR}, 2020.

\bibitem{suris2019learning}
D.~Suris, A.~Recasens, D.~Bau, D.~Harwath, J.~Glass, and A.~Torralba,
  ``Learning words by drawing images,'' in \emph{CVPR}, 2019.

\bibitem{mortazavi2020speech}
M.~S. Mortazavi, ``Speech-image semantic alignment does not depend on any prior
  classification tasks,'' \emph{INTERSPEECH}, 2020.

\bibitem{sanabria2021talk}
R.~Sanabria, A.~Waters, and J.~Baldridge, ``Talk, don't write: A study of
  direct speech-based image retrieval,'' \emph{arXiv preprint
  arXiv:2104.01894}, 2021.

\bibitem{wang2021align}
L.~Wang, X.~Wang, M.~Hasegawa-Johnson, O.~Scharenborg, and N.~Dehak, ``Align or
  attend? toward more efficient and accurate spoken word discovery using
  speech-to-image retrieval,'' in \emph{ICASSP}, 2021.

\bibitem{harwath2017learning}
D.~Harwath and J.~Glass, ``Learning word-like units from joint audio-visual
  analysis,'' in \emph{ACL}, 2017.

\bibitem{zhou2014learning}
B.~Zhou, A.~Lapedriza, J.~Xiao, A.~Torralba, and A.~Oliva, ``Learning deep
  features for scene recognition using places database,'' in \emph{NeurIPS},
  2014.

\bibitem{chrupala2021visually}
G.~Chrupa{\l}a, ``Visually grounded models of spoken language: A survey of
  datasets, architectures and evaluation techniques,'' \emph{arXiv preprint
  arXiv:2104.13225}, 2021.

\bibitem{monfort2021spoken}
M.~Monfort, S.~Jin, A.~Liu, D.~Harwath, R.~Feris, J.~Glass, and A.~Oliva,
  ``Spoken moments: Learning joint audio-visual representations from video
  descriptions,'' in \emph{CVPR}, 2021.

\bibitem{oncescu2020queryd}
A.-M. Oncescu, J.~F. Henriques, Y.~Liu, A.~Zisserman, and S.~Albanie, ``Queryd:
  A video dataset with high-quality textual and audio narrations,'' \emph{arXiv
  preprint arXiv:2011.11071}, 2020.

\bibitem{holzenberger2019learning}
N.~Holzenberger, S.~Palaskar, P.~Madhyastha, F.~Metze, and R.~Arora, ``Learning
  from multiview correlations in open-domain videos,'' in \emph{ICASSP}, 2019.

\bibitem{wray2019fine}
M.~Wray, D.~Larlus, G.~Csurka, and D.~Damen, ``Fine-grained action retrieval
  through multiple parts-of-speech embeddings,'' in \emph{ICCV}, 2019.

\bibitem{alayrac2020self}
J.-B. Alayrac, A.~Recasens, R.~Schneider, R.~Arandjelovi{\'c}, J.~Ramapuram,
  J.~De~Fauw, L.~Smaira, S.~Dieleman, and A.~Zisserman, ``Self-supervised
  multimodal versatile networks,'' \emph{NeurIPS}, vol.~33, 2020.

\bibitem{chen2021multimodal}
B.~Chen, A.~Rouditchenko, K.~Duarte, H.~Kuehne, S.~Thomas, A.~Boggust,
  R.~Panda, B.~Kingsbury, R.~Feris, D.~Harwath \emph{et~al.}, ``Multimodal
  clustering networks for self-supervised learning from unlabeled videos,''
  \emph{arXiv preprint arXiv:2104.12671}, 2021.

\bibitem{sanabria18how2}
R.~Sanabria, O.~Caglayan, S.~Palaskar, D.~Elliott, L.~Barrault, L.~Specia, and
  F.~Metze, ``{How2:} a large-scale dataset for multimodal language
  understanding,'' in \emph{Workshop on Visually Grounded Interaction and
  Language (ViGIL)}.\hskip 1em plus 0.5em minus 0.4em\relax NeurIPS, 2018.

\bibitem{boggust2019grounding}
A.~Boggust, K.~Audhkhasi, D.~Joshi, D.~Harwath, S.~Thomas, R.~Feris,
  D.~Gutfreund, Y.~Zhang, A.~Torralba, M.~Picheny, and J.~Glass, ``Grounding
  spoken words in unlabeled video,'' in \emph{CVPR Sight and Sound Workshop},
  2019.

\bibitem{kamper2018visually}
H.~Kamper and M.~Roth, ``Visually grounded cross-lingual keyword spotting in
  speech,'' \emph{arXiv preprint arXiv:1806.05030}, 2018.

\bibitem{azuh2019towards}
E.~Azuh, D.~Harwath, and J.~R. Glass, ``Towards bilingual lexicon discovery
  from visually grounded speech audio.'' in \emph{INTERSPEECH}, 2019.

\bibitem{ohishi2020pair}
Y.~Ohishi, A.~Kimura, T.~Kawanishi, K.~Kashino, D.~Harwath, and J.~Glass,
  ``Pair expansion for learning multilingual semantic embeddings using disjoint
  visually-grounded speech audio datasets,'' \emph{INTERSPEECH}, 2020.

\bibitem{wang2019vatex}
X.~Wang, J.~Wu, J.~Chen, L.~Li, Y.-F. Wang, and W.~Y. Wang, ``Vatex: A
  large-scale, high-quality multilingual dataset for video-and-language
  research,'' in \emph{ICCV}, 2019.

\bibitem{sigurdsson2020visual}
G.~A. Sigurdsson, J.-B. Alayrac, A.~Nematzadeh, L.~Smaira, M.~Malinowski,
  J.~Carreira, P.~Blunsom, and A.~Zisserman, ``Visual grounding in video for
  unsupervised word translation,'' in \emph{CVPR}, 2020.

\bibitem{huang2013cross}
J.-T. Huang, J.~Li, D.~Yu, L.~Deng, and Y.~Gong, ``Cross-language knowledge
  transfer using multilingual deep neural network with shared hidden layers,''
  in \emph{ICASSP}, 2013.

\bibitem{karafidt2018analysis}
M.~Kar{\'a}fidt, M.~K. Baskar, K.~Vesel{\`y}, F.~Gr{\'e}zl, L.~Burget, and
  J.~{\v{C}}ernock{\`y}, ``Analysis of multilingual blstm acoustic model on low
  and high resource languages,'' in \emph{ICASSP}, 2018.

\bibitem{cho2018multilingual}
J.~Cho, M.~K. Baskar, R.~Li, M.~Wiesner, S.~H. Mallidi, N.~Yalta, M.~Karafiat,
  S.~Watanabe, and T.~Hori, ``Multilingual sequence-to-sequence speech
  recognition: architecture, transfer learning, and language modeling,'' in
  \emph{SLT}, 2018.

\bibitem{conneau2020unsupervised}
A.~Conneau, A.~Baevski, R.~Collobert, A.~Mohamed, and M.~Auli, ``Unsupervised
  cross-lingual representation learning for speech recognition,'' \emph{arXiv
  preprint arXiv:2006.13979}, 2020.

\bibitem{thomas2012multilingual}
S.~Thomas, S.~Ganapathy, and H.~Hermansky, ``Multilingual mlp features for
  low-resource lvcsr systems,'' in \emph{ICASSP}, 2012.

\bibitem{ghoshal2013multilingual}
A.~Ghoshal, P.~Swietojanski, and S.~Renals, ``Multilingual training of deep
  neural networks,'' in \emph{ICASSP}, 2013.

\bibitem{ddoukhanicassp2018}
D.~Doukhan, J.~Carrive, F.~Vallet, A.~Larcher, and S.~Meignier, ``An
  open-source speaker gender detection framework for monitoring gender
  equality,'' in \emph{ICASSP}, 2018.

\bibitem{he2016deep}
K.~He, X.~Zhang, S.~Ren, and J.~Sun, ``Deep residual learning for image
  recognition,'' in \emph{CVPR}, 2016.

\bibitem{hara2018can}
K.~Hara, H.~Kataoka, and Y.~Satoh, ``Can spatiotemporal 3d cnns retrace the
  history of 2d cnns and imagenet?'' in \emph{CVPR}, 2018.

\end{thebibliography}
\end{document}